\documentclass[letterpaper, 10 pt, conference]{ieeeconf}  

\usepackage[utf8]{inputenc}
\usepackage[T1]{fontenc}
\usepackage{mathptmx} 
\usepackage{mathptmx} 
\usepackage{amsmath} 
\usepackage{amssymb}  
\usepackage{graphicx} 
\usepackage{grffile}
\usepackage{array}
\usepackage{xcolor}
\usepackage{mathtools}
\usepackage{commath}
\usepackage{colortbl}
\usepackage{enumerate}
\usepackage{hyperref}
\usepackage{float}
\usepackage{algorithmic}
\usepackage{balance}
\usepackage{nopageno} 

\title{\LARGE \bf
Demonstrating a Robust Walking Algorithm for Underactuated Bipedal Robots in Non-flat, Non-stationary Environments}

\author{Oluwami Dosunmu-Ogunbi, Aayushi Shrivastava,
Jessy W Grizzle}

\begin{document}
\maketitle
\thispagestyle{plain}
\pagestyle{plain}

\begin{abstract}\label{sec:abstract} 
This work explores an innovative algorithm designed to enhance the mobility of underactuated bipedal robots across challenging terrains, especially when navigating through spaces with constrained opportunities for foot support, like steps or stairs. By combining ankle torque with a refined angular momentum-based linear inverted pendulum model (ALIP), our method allows variability in the robot's center of mass height. We employ a dual-strategy controller that merges virtual constraints for precise motion regulation across essential degrees of freedom with an ALIP-centric model predictive control (MPC) framework, aimed at enforcing gait stability. The effectiveness of our feedback design is demonstrated through its application on the Cassie bipedal robot, which features 20 degrees of freedom. Key to our implementation is the development of tailored nominal trajectories and an optimized MPC that reduces the execution time to under 500 microseconds—and, hence, is compatible with Cassie's controller update frequency. This paper not only showcases the successful hardware deployment but also demonstrates a new capability, a bipedal robot using a moving walkway.
\end{abstract}


\section{Introduction} \label{sec:introduction}
Bipedal robots exhibit substantial potential for aiding humans across various sectors, including hospitals, disaster management, factories, and construction sites. However, these environments often present challenges with constrained footholds, impeding conventional foot placement techniques due to limited space available for robot stepping.

Our approach introduces a dynamic variation that enables the robot's center of mass height to adapt within each step, addressing the limitations posed by confined footholds. We first introduce our methodology and present simulation results in the SimMechanics environment in \cite{dosunmu2023stair}. While these preliminary results are commendable, a significant challenge lies in transitioning from simulation to real-world applications. Bridging this gap is hindered by persistent difficulties in accounting for factors such as motor friction, sensor inaccuracies, and unforeseen variables that are not considered in simulation environments. Notably, deploying controllers on Cassie hardware poses a critical challenge due to the stringent maximum controller execution time of 0.5 milliseconds. The nominal model predictive controller (MPC) responsible for computing ankle torque demands an execution time exceeding this limit. In this paper, we discuss how we overcome these challenges and successfully implement our novel control method on the Cassie bipedal robot hardware.


\subsection{Background}
Overly specific models can be computationally burdensome and lack versatility across robotic systems. Simplified models enable diverse control strategies, but may not capture the primary dynamics effectively, limiting system agility. Bridging the sim-to-real gap remains challenging. We build on the Angular Momentum Linear Inverted Pendulum (ALIP) model from \cite{powell2016, gong2021}, enhancing it to better accommodate changes in center of mass (CoM) height, as detailed in \cite{dosunmu2023stair}.

Several scholars, including Fu et al. \cite{Fu2008} and Caron et al. \cite{Caron2019}, have explored the development of controllers tailored for fully actuated humanoid robots with 32 and 34 Degrees of Freedom (DoF) respectively, focusing on stair navigation. In \cite{hereid2019}, the authors devised open-loop stair gaits for the 3D underactuated 20 DoF Cassie bipedal robot, omitting investigation into closed-loop control.
In \cite{verhagen2022}, researchers managed to simulate human-like movements during planned and unplanned descents on the Cassie bipedal robot. Our paper aims to enhance the navigational capabilities of the Cassie biped by achieving an asymptotically stable periodic gait on non-flat, non-stationary terrain.

Previous research by Siekmann et al. \cite{Siekmann2021} employed reinforcement learning to devise a closed-loop controller for the Cassie bipedal robot, treating changes in terrain elevation as an unpredictable perturbation. However, this approach resulted in significant impacts, posing potential risks to the robot's structural integrity. The authors of \cite{Nguyen2016} also explore dynamic walking on constrained footholds on level ground using control barrier functions. Dai et al. \cite{Dai2022} tackled similar challenges by developing a dynamic walking controller for non-flat constrained footholds, including staircases, by regulating the underactuated robot’s vertical Center of Mass (CoM). We implement an alternative approach to navigation in varied terrain, utilizing the often-overlooked stance ankle motor alongside virtual constraint-based control. 

The authors of \cite{acosta2023bipedal} integrate the ankle torque model in their application of a terrain-aware controller also capable of walking on constrained footholds with successful implementation on hardware. In \cite{xiong2021slip}, the authors developed a novel controller to enable navigation over rough and challenging terrains for the actuated Spring Loaded Inverted Pendulum Model. This paper seeks to prove the validity of our novel robust controller on hardware to extend on initial SimMechanics results.

While the utilization of virtual constraints in legged locomotion is not novel \cite{Griffin_Grizzle_2017,gong2021zero,gibson2022terrain,gong2019feedback,Hereid_2016_HZD_DirectCollocation,Sreenath_2011_HZD_Mabel_Walking,Hamed_Kim_Pandala_2020}, applying them to a non-flat environment for bipedal robot navigation, as we do in this work, presents an innovative application.

\subsection{Contributions}
\begin{enumerate}

\item Presentation of the nominal trajectories tailored for various inclined planes.
\item Implementation of a foot placement strategy for lateral stabilization of the robot that can be implemented in real-time.
\item Enhancement of the Model Predictive Controller's efficiency in computing ankle torque for the sagittal plane. This involves linearizing impacts around nominal trajectories and the strategic offloading of computations to a secondary computer and the utilization of UDP communication between the secondary computer and Cassie.
\item Validation of the full controller on the Cassie bipedal robot hardware.

\end{enumerate}


\section{Review of the Control Philosophy}
\label{sec:Philosophy}
In this section, we review the control philosophy outlined in our previous work \cite{dosunmu2023stair} and elucidate the enhancements made to ensure its effectiveness in conducting hardware experiments.

The Cassie biped has 20 DoF to control. During single support, where one foot is grounded and the other is free, nine DoF are subject to holonomic constraints: four from Cassie's springs (two per leg) and five from the stance foot. Thus, we are left with 11 DoF to control and ten actuators. The robot is therefore underactuated.

We employ Passivity-Based Control (PBC) to regulate an additional nine DoF, utilizing nine of Cassie's actuators to track nominal trajectories generated offline. The tenth actuator is assigned to regulate one DoF, specifically the stance toe actuator, through a model predictive controller (MPC). The last DoF is managed via a lateral foot placement approach.

To facilitate the successful execution of the hardware experiments described in this paper, several critical adjustments were made. Firstly, new nominal trajectories were generated to align with the new experimental scenarios discussed in this study, as discussed in detail in Section \ref{sec:nominal}. Subsequently, enhancements were strategically implemented to enhance the efficiency and precision of the MPC algorithm. This involved transferring its computations to a secondary computer, integrating a real-time nonlinear optimization tool, and devising a new impact map based on linearization around the nominal trajectories, as elucidated in Section \ref{sec:sim2real}. Finally, a real-time method was devised to compute the lateral foot placement strategy, detailed in Section \ref{sec:lateralControl}.

\section{Nominal Periodic Orbit}
\label{sec:nominal}

We utilized Fast Robot Optimization and Simulation Toolkit (FROST) \cite{Hereid2017FROST} to generate nominal trajectories for Cassie, leveraging its advanced framework designed for optimizing and simulating the full-body dynamics of bipedal walking robots. FROST seamlessly integrates virtual constraints-based feedback controllers and employs a Wolfram Mathematica backend to symbolically generate expressions for multi-domain system dynamics and kinematics. These symbolic expressions are then translated into C/C++ code, compiled into *.MEX files under MATLAB, enhancing computational speed critical for hardware implementation.

FROST conceptualizes dynamic bipedal walking as a hybrid system, combining both continuous phases and discrete transitions, represented by a Directed Graph. A notable feature is its use of state-of-the-art direct collocation approaches for gait optimization, ensuring swift and dependable convergence. The default control law of FROST relies on virtual constraints-based feedback controllers. This toolkit emerges as a versatile and efficient resource for researchers exploring bipedal robot control and dynamics optimization.

In our study, FROST was employed to generate multiple nominal trajectories for Cassie. We generated trajectories for various scenarios, including marching in place, walking forward, and transitioning from walking on flat ground to different inclines (4 degrees, 8 degrees, 15 degrees, and 20 degrees). Utilizing B\`ezier curves \cite{farin1983algorithms}, we approximated optimized trajectories for the nine virtual constraints (defined in \cite{dosunmu2023stair}), as well as the angular momentum and CoM angle crucial for MPC computation for ankle torque, as detailed in \cite{dosunmu2023stair}.

\subsection{Generating Nominal Trajectories with FROST}
\begin{figure}
    \centering
    \includegraphics[width=0.45\textwidth]{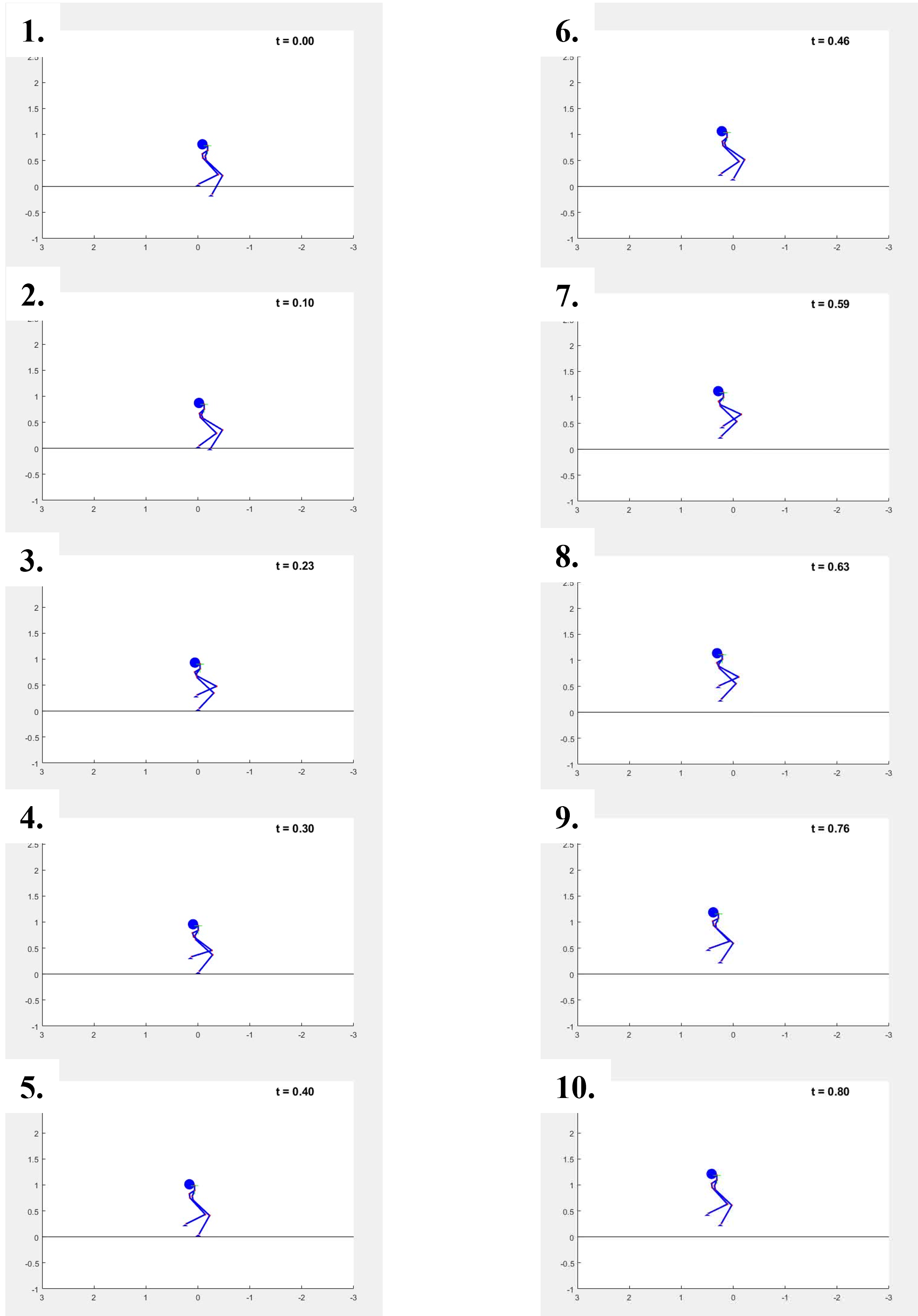}
    \caption{Outtakes of a nominal trajectory for a non-flat terrain generated by FROST using a full-order model of the Cassie biped. The FROST animation is restricted to only showing a flat line at $y$=0. However, one can see that a non-flat trajectory is being followed by observing how the feet are placed with respect to the $y$=0 line.}
    \label{fig:FROST_stairs}
\end{figure}

Figure \ref{fig:FROST_stairs} illustrates animation outtakes from a nominal trajectory for non-flat terrain designed using FROST and a full-order model of the Cassie biped. The trajectory encompasses two steps, accommodating both right and left foot stance phases. The generation of this trajectory involved addressing several key constraints used for all the nominal trajectories.

A periodicity constraint was imposed at the conclusion of the second step, enabling the use of this trajectory to traverse an unlimited number of steps through the simple repetition of the nominal trajectory. Another constraint was incorporated to minimize torque on the stance ankle motor. This strategic addition provides increased adaptability when applying the trajectory to an actual Cassie robot. Our hypothesis posited that a nominal trajectory minimizing stance ankle torque would translate to minimal ankle torque requirements for the physical robot when walking unperturbed, allowing more room to overcome disturbances by utilizing stance ankle torque output. This hypothesis was validated, as discussed in Section \ref{section:Hardware}.

\begin{figure}
    \centering
    \includegraphics[width=0.45\textwidth]{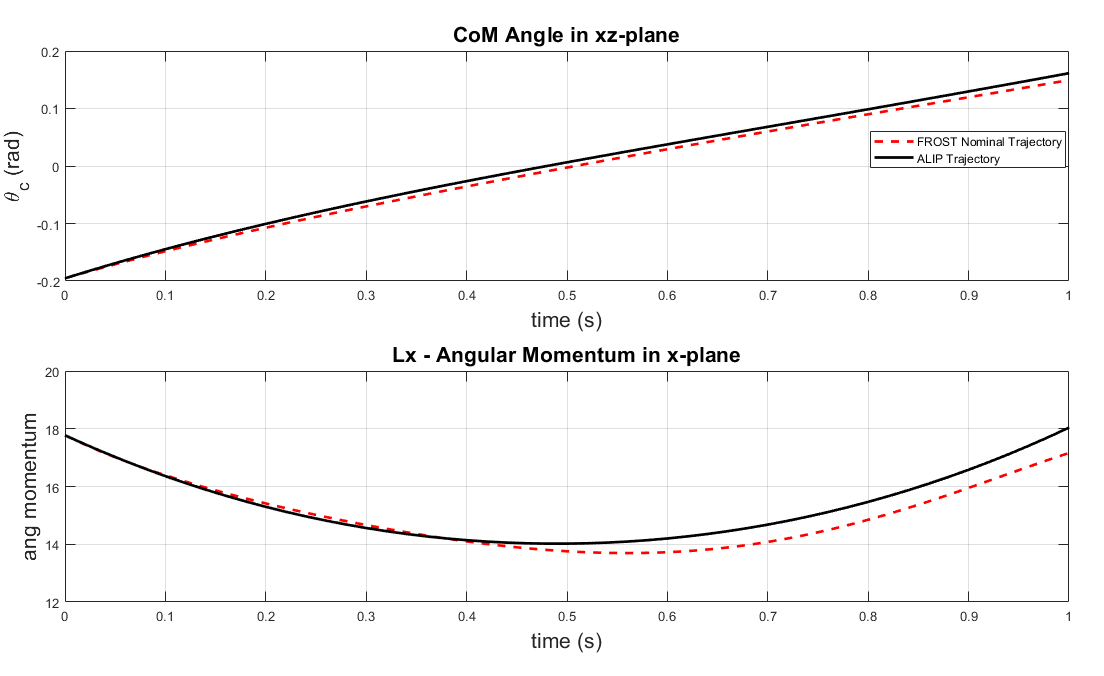}
    \caption{Comparison of the nominal trajectory generated by FROST and ALIP model of angular momentum and CoM angle.}
    \label{fig:CompareFROST_ALIP}
\end{figure}

In Figure \ref{fig:CompareFROST_ALIP}, we present plots depicting the angular momentum and CoM angle, comparing the results computed by FROST for the nominal trajectories against the values obtained by the new variation of the ALIP model. As anticipated, the values are similar. This shows that the trajectories generated by FROST satisfy the ALIP model.

\subsection{Bézier Curves}
We translate the numerical values of the nominal trajectories generated by FROST and express them in the language of B\'ezier curves so that they can be more readily used and manipulated by our controller. A B\'ezier curve is a mathematical representation of a curve defined by two or more ``control points,'' which may be situated either on the curve itself or externally \cite{westerveltfeedback}.

\begin{figure}
    \centering
    \includegraphics[width=0.45\textwidth]{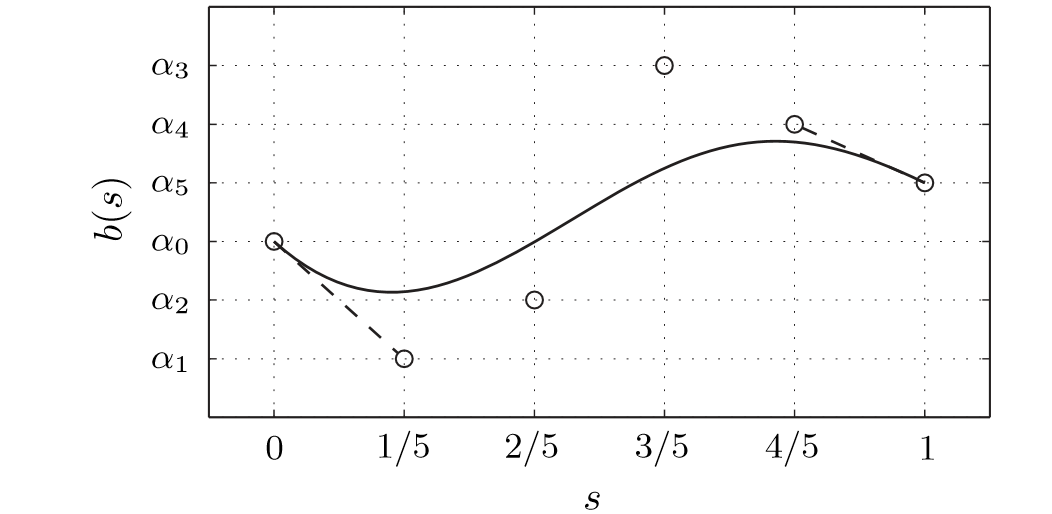}
    \caption{Graph of a B\'ezier Curve of Order 5 \cite{westerveltfeedback}.}
    \label{fig:bezcurve5}
\end{figure}

Figure \ref{fig:bezcurve5} illustrates a B\'ezier curve of order $5$, featuring six control points (the order plus one). The curve lies within the convex hull of these control points. Notably, the curve commences at $b(0) = \alpha_0$ and concludes at $b(1) = \alpha_5$. This alignment is intentional, as for all B\'ezier curves, $b_i(0) = \alpha_0^i$ and $b_i(1) = \alpha_M^i$, signifying that the $i^{\text{th}}$ B\'ezier curve starts at the first coefficient $\alpha_0^i$ and concludes at the last coefficient $\alpha_M^i$.

By incorporating B\'ezier curves into our controller for representing the nominal trajectories from FROST, we gain the ability to finely manipulate these trajectories through adjustments to their control points. For instance, to extend the time the swing foot takes to ascend and avoid contact with an obstacle (e.g., a stair), we strategically reposition the interior control points associated with the sagittal plane motion of the swing foot, moving them closer to the initial control point. This adjustment effectively prolongs the swing foot's retention of its sagittal position as it elevates in the $z-$ plane. The utilization of B\'ezier curves empowers us to make nuanced refinements to the nominal trajectories, such as this, without the need for the time-intensive task of regenerating trajectories through offline optimization.

\section{Lateral Stabilization of the Robot}
\label{sec:lateralControl}
We stabilized the lateral motion of the Cassie biped by using the angular momentum-based foot placement strategy developed in \cite{gong2021}, but with the new ALIP model derived in our work \cite{dosunmu2023stair}. This section derives the foot placement strategy for the new ALIP model. 

Unlike the previous version of the ALIP model, the new ALIP model does not have a closed-form solution. Thus, a numerical approach must be implemented for a foot placement strategy. To ensure that this approach can be implemented on hardware, we developed a look-up table to be used in real-time applications. This look-up table approach renders the execution time for the lateral foot placement controller to be less than 5 $\mu$s, critical to ensure that we adhere to Cassie's stringent 2 kHz control-update frequency.

The strategy to compute lateral foot placement numerically is outlined as follows:
\begin{enumerate}
    \item Use Euler method for numerical integration to estimate the angular momentum at the end of the \textbf{next} step, \textbf{before} impact, for two lateral foot positions $y_{\text{st}\rightarrow\text{sw}}^1$ and $y_{\text{st}\rightarrow\text{sw}}^2$, where $y_{\text{st}\rightarrow\text{sw}}^1$ and $y_{\text{st}\rightarrow\text{sw}}^2$ are close in value. We will call these two angular momentum values $L_1$ and $L_2$. We pick the first lateral position $y_{\text{st}\rightarrow\text{sw}}^1$ based off of the nominal trajectory and the second lateral position $y_{\text{st}\rightarrow\text{sw}}^2$ to be a value close to the first.
    \item Define a desired angular momentum $L_{\text{des}}$ that we assume to be approximately along the line between Point 1 ($y_{\text{st}\rightarrow\text{sw}}^1$, $L_1$) and Point 2 ($y_{\text{st}\rightarrow\text{sw}}^2$, $L_2$).
    \item Using 2D or 3D linear interpolation, find $y_{\text{des}}$, the lateral foot placement position corresponding to $L_\text{des}$.
\end{enumerate}

\subsection{Euler Integration}
Euler integration, also known as the Euler method, is a numerical technique for approximating the solution of an Ordinary Differential Equation (ODE) with an initial value problem \cite{biswas2013discussion}. The Euler method proceeds by discretizing the time domain into small steps and approximating the solution at each step. The iterative formula for Euler integration is:
\begin{equation}
    x_{n+1} = x_n + \Delta t \cdot f(t_n,x_n)
    \label{eq:eulerIntegration}
\end{equation}
where $x_{n+1}$ is the approximation of the solution at the next time step, $x_n$ is the solution at the current time step, $\Delta t$ is the step size representing the time interval between consecutive steps, $f(t_n,x_n)$ is a given first-order ODE function evaluated at time $n$, and $t_n$ is the current time.

The method starts with an initial value $x_0$ and $t_0$, and subsequent values are computed using the iterative formula. The smaller the step size $\Delta t$, the more accurate the approximation, but it also increases the computational cost.

In our application, we determine our initial values $x_0$ and $t_0$ as being the current state of the system $x_0 = \left[ \theta_c(t_0) ~~ L(t_0) \right]^\top$ at time $t_0$. We use Equation \eqref{eq:eulerIntegration} to compute the state of the system at every time step $\Delta t$ until the end of the \textbf{first} step of a period length $t = T$, \textbf{before} impact, where $f(t_n,x_n)$ is defined as the system dynamics from the new ALIP model from \cite{dosunmu2023stair}. We use an impact map (derived in Section \ref{sec:impactARIP}) to compute the system dynamics \textbf{after} impact of the \textbf{first} step. These state values are equivalent to the state values at the beginning of the \textbf{next} step. Using these post-impact values as initial values, we then implement Euler integration again to compute the state of the system at every time step from the beginning of the \textbf{next} step until the end of the step \textbf{before} impact.

\subsection{Impact Map for the New ALIP Model}
\label{sec:impactARIP}
To compute the state of the system \textbf{after} impact of the \textbf{first} step, we develop an impact map for the new ALIP model. This new impact model for the reduced order model is separate from the impact map for the full-order Cassie model. 

\begin{figure}
    \centering
    \includegraphics[width=0.45\textwidth]{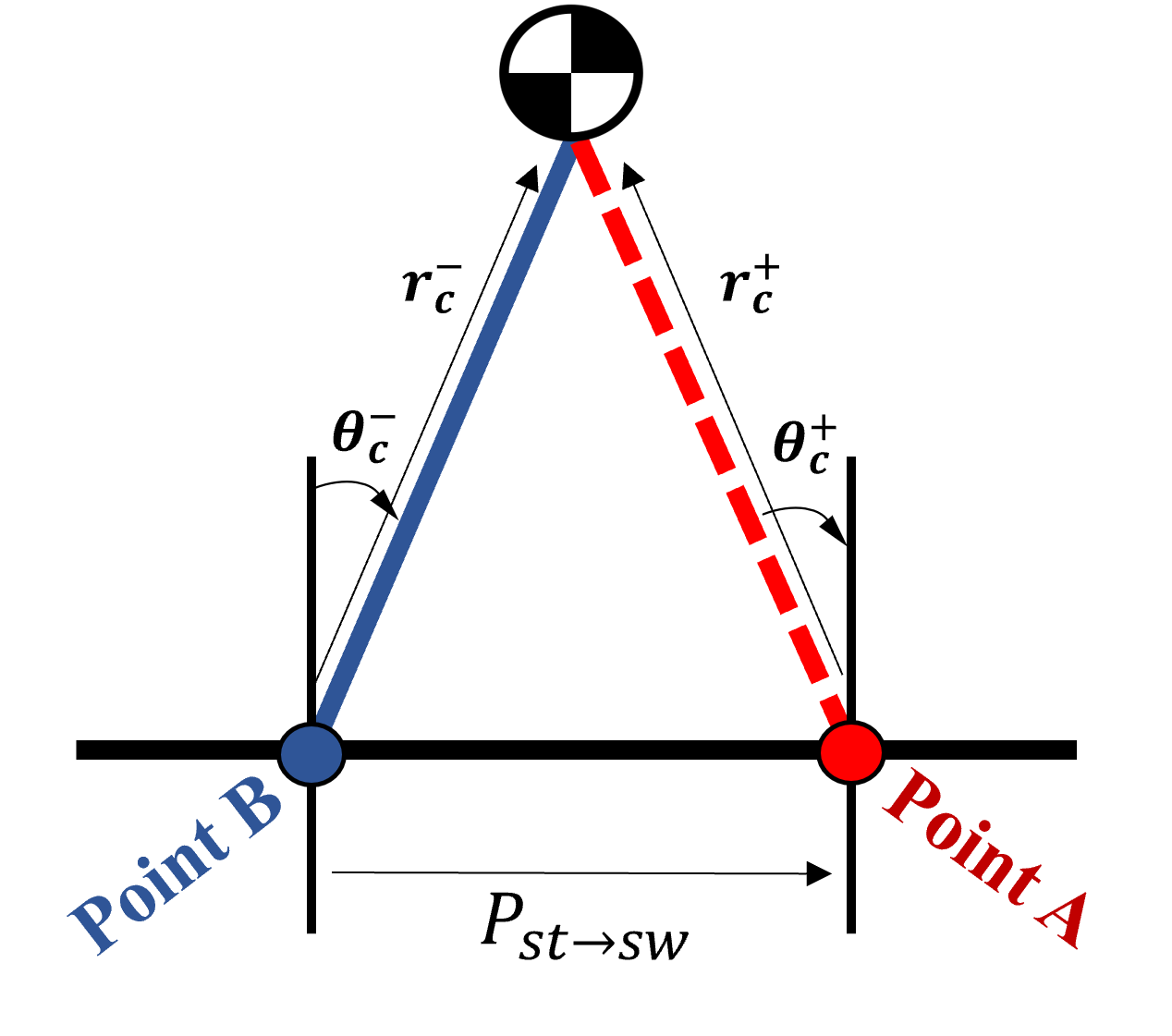}
    \caption{Schematic of the inverted pendulum to derive the impact map for the new variation on the ALIP model.}
    \label{fig:ARIPimpact}
\end{figure}

Figure \ref{fig:ARIPimpact} illustrates the relevant variables to derive the new impact map. 
Let Point A be the position where the swing foot touches the ground and Point B be the point where the stance foot touches the ground. Using the angular momentum about Point B before impact $L_B^-$, we can compute the angular momentum about Point A after impact $L_A^+$. From the conservation of angular momentum, we know that the angular momentum about Point A before impact $L_A^-$ is equivalent to the angular momentum about Point A after impact $L_A^+$. That is,
\begin{equation}
    L_A^- = L_A^+.
    \label{eq:conservAngMom}
\end{equation}

Using the \textit{angular momentum transfer formula} \cite{gong2021zero}, we compute $L_B^-$ from $L_A^-$,
\begin{equation}
    L_{A}^- = L_B^- + P_\text{st $\rightarrow$ sw} \times m v_c
    \label{eq:angMomTransFunc}
\end{equation}
where $P_\text{st $\rightarrow$ sw} = \left[ P_\text{st $\rightarrow$ sw}^x ~~ P_\text{st $\rightarrow$ sw}^z \right]^\top$ is the vector from the stance to swing foot, $m$ is the mass of the robot, and $v_c$ is the CoM velocity. In the x-z plane, the position of the CoM is defined as
\begin{equation}
    p_c = r_c^- 
    \begin{bmatrix}
        \sin(\theta_c^-)\\
        \cos(\theta_c^-)
    \end{bmatrix}.
    \label{eq:pcom}
\end{equation}
Taking the derivative Equation \eqref{eq:pcom} yields the velocity of the CoM,
\begin{equation}
    \frac{d}{dt}p_c = v_c = 
    \begin{bmatrix}
        r_c^- \cos(\theta_c^-) \cdot \dot{\theta}_c^- + \dot{r}_c^- \sin(\theta_c^-) \\
        -r_c^- \sin(\theta_c^-) \cdot \dot{\theta}_c^- + \dot{r}_c^- \cos(\theta_c^-)
    \end{bmatrix}.
    \label{eq:vcomARIP}
\end{equation}

From Equation \eqref{eq:conservAngMom}, Equation \eqref{eq:angMomTransFunc}, and Equation \eqref{eq:vcomARIP}, we compute
\small
\begin{equation}
    \begin{aligned}
        L_{A}^- = L_A^+ = L_B^- + m \Bigg[P_\text{st $\rightarrow$ sw}^z \Big(r_c^- \cos (\theta_c^-) \dot{\theta}_c^- + \dot{r}_c^- \sin(\theta_c^-) \Big) - \\
        P_\text{st $\rightarrow$ sw}^x \Big(-r_c^- \sin (\theta_c^-) \dot{\theta}_c^- + \dot{r}_c^- \cos(\theta_c^-) \Big) \Bigg] .
    \end{aligned}
    \label{eq:LAplus}
\end{equation}
\normalsize
Using trigonmetry, we determine the value of the CoM angle after impact $\theta_c^+$ to be
\begin{equation}
    \theta_c^+ = \arccos \Bigg( \frac{r_c^- \cos \theta_c^- - P_\text{st $\rightarrow$ sw}^z}{r_c^+} \Bigg).
    \label{eq:thetacplus}
\end{equation}

Equation \eqref{eq:LAplus} and Equation \eqref{eq:thetacplus} form our impact map. That is, the state of the system after impact is defined as
\small
\begin{equation}
    x^+ = 
    \begin{bmatrix}
        \theta_c^+ \\
        L^+
    \end{bmatrix} =
    \begin{bmatrix}
        \arccos \Bigg( \frac{r_c^- \cos \theta_c^- - P_\text{st $\rightarrow$ sw}^z}{r_c^+} \Bigg) \\
        \begin{matrix}
            L_B^- + m \Bigg[P_\text{st $\rightarrow$ sw}^z \Big(r_c^- \cos (\theta_c^-) \dot{\theta}_c^- + \dot{r}_c^- \sin(\theta_c^-) \Big)\\
           - P_\text{st $\rightarrow$ sw}^x \Big(-r_c^- \sin (\theta_c^-) \dot{\theta}_c^- + \dot{r}_c^- \cos(\theta_c^-) \Big) \Bigg] 
        \end{matrix}
    \end{bmatrix}.
\end{equation}
\normalsize
\subsection{Linear Interpolation}
We can compute the desired lateral foot placement 
$y_\text{des}$ by

\begin{equation}
    y_\text{des} = y_{\text{st}\rightarrow\text{sw}}^1 + \frac{L_\text{des}-L_1}{\text{slope}},
\end{equation}

where
\begin{equation}
    \text{slope} = \frac{L_2 - L_1}{y_{\text{st}\rightarrow\text{sw}}^2 - y_{\text{st}\rightarrow\text{sw}}^1}.
\end{equation}


Similarly, one could use CoM angle, $\theta$, instead of angular momentum to compute lateral foot placement. That is,
\begin{equation}
    y_\text{des} = y_{\text{st}\rightarrow\text{sw}}^1 + \frac{\theta_\text{des}-\theta_1}{\text{slope}},
\end{equation}
where
\begin{equation}
    \text{slope} = \frac{\theta_2 - \theta_1}{y_{\text{st}\rightarrow\text{sw}}^2 - y_{\text{st}\rightarrow\text{sw}}^1}.
\end{equation}


\section{Navigating the sim-to-real Gap for Cassie Hardware}
\label{sec:sim2real}
A paramount concern in deploying controllers on Cassie hardware is adhering to the stringent maximum controller execution time of 2 kHz. The passivity-based controller (PBC) detailed in \cite{dosunmu2023stair} adeptly operates within this temporal constraint. In contrast, the model predictive controller (MPC) also discussed in \cite{dosunmu2023stair} demands a higher execution time. To surmount this challenge, we strategically offloaded the MPC computations to a secondary computer. The execution time was further optimized through the utilization of CasADi, an open-source tool for nonlinear optimization and algorithmic differentiation \cite{Andersson_2019_Casadi,ANDERSSON2018_qrqp}. The MPC computation now runs at an approximate duration of 500 $\mu$s. While it was not strictly necessary to further reduce the MPC computation time when we offloaded it to a secondary computer, it is nevertheless good practice to optimize the code whenever possible. Given that it is unnecessary to compute a desired ankle torque value for every time step, the utilization of a secondary computer for MPC, with a slower execution time dedicated to ankle torque values, poses no hindrance. This claim is substantiated by the positive outcomes observed in SimMechanics simulation experiments, where the transfer of ankle torque computations to the secondary computer did not result in any noticeable deviations in results.

To address the observed spikes in ankle torque values during the SimMechanics simulation, as discussed in \cite{dosunmu2023stair}, we developed an improved impact map based on the linearization of the full-order impact map about the nominal trajectories. This enhancement reduced unexpected spikes in control output, further enhancing the stability and performance of Cassie in the hardware experiments discussed in this paper.

\section{Hardware Results and Discussion} \label{section:Hardware}


This section delves into the practical implementation of the controllers and strategies outlined in \cite{dosunmu2023stair} and Sections \ref{sec:nominal}, \ref{sec:lateralControl}, and \ref{sec:sim2real} on the physical 20 DOF Cassie bipedal robot hardware. While the gait demonstrations of \cite{dosunmu2023stair} in the SimMechanics simulation environment were promising, the pivotal challenge lies in bridging the gap between simulation and reality. 
\subsection{Maximum Walking Speed on Various Inclined Surfaces}
\label{sec:hardwareMaxWalkingSpeed}
In evaluating the performance of Cassie equipped with the novel controller structure developed in this paper, a series of tests were conducted to determine its maximum walking speed on diverse inclinations. Previous walking controllers for Cassie hardware demonstrated speeds of 2 m/s on level ground \cite{gong2021zero}, but exploration of maximum speeds on inclined surfaces was lacking.

With the implementation of our new controller, Cassie showcased an enhanced walking capability, achieving a comfortable speed of up to 2.2 m/s on a flat treadmill \cite{DynamicLegLocomotion_22Flat}. Extending the analysis to inclined surfaces, Cassie maintained a speed of 2 m/s on a treadmill inclined at 4 degrees \cite{DynamicLegLocomotion_FlatTo4Deg} and 8 degrees \cite{DynamicLegLocomotion_FlatTo8Deg}. Even on steeper inclinations, such as a 15- and 20-degree slope, Cassie exhibited a walking speed, reaching up to 1.5 m/s \cite{DynamicLegLocomotion_FlatTo15Deg,DynamicLegLocomotion_FlatTo20Deg}. These findings underscore the adaptability and improved performance achieved through the integration of the developed controller structure, opening avenues for enhanced mobility on surfaces with varying degrees of incline.
\subsection{Continuous Walking on Changing Incline}

In our subsequent experiment, we conducted a continuous walking test on a dynamically changing incline. The experiment commenced on a level surface with a 0-degree incline, progressively transitioning to the maximum incline of 20 degrees, and then returning to the initial 0-degree incline, as depicted in \cite{DynamicLegLocomotion_ContinuousInclineWalking}. Throughout the experiment, the treadmill maintained a constant speed of 0.9 m/s. To streamline the complexity of the experiment, we employed only three nominal trajectories:

\begin{itemize}
\item \textbf{Nominal Trajectory for Flat Ground Walking:}
Implemented during Cassie's movement on inclines ranging from 0 to 7 degrees.

\item \textbf{Nominal Trajectory for 8-Degree Incline:}
Applied as Cassie traversed inclines ranging from 8 to 14 degrees.

\item \textbf{Nominal Trajectory for 15-Degree Incline:}
Utilized when Cassie navigated inclines ranging from 15 to 20 degrees.
\end{itemize}

The use of nominal trajectories on inclines for which the trajectories were not specifically optimized showcased the robustness of the controller. This experiment demonstrated that perfect trajectories for all situations are not a prerequisite for stability—--for example, a nominal trajectory optimized for a 15-degree incline successfully operated on a 20-degree inclined surface.

Moreover, all nominal trajectories assumed a constant nominal speed of 0.5 m/s, introducing additional disturbance to the system since the treadmill operated at 0.9 m/s during the experiment. This underscored the controller's ability to maintain stability and adjust seamlessly to varying walking speeds, irrespective of the speed specified in the nominal trajectory. The experimental configuration provided a comprehensive assessment of the developed controller's capacity to adapt and ensure stability during continuous walking on inclines.

\begin{figure}
    \centering
    \includegraphics[width=0.45\textwidth]{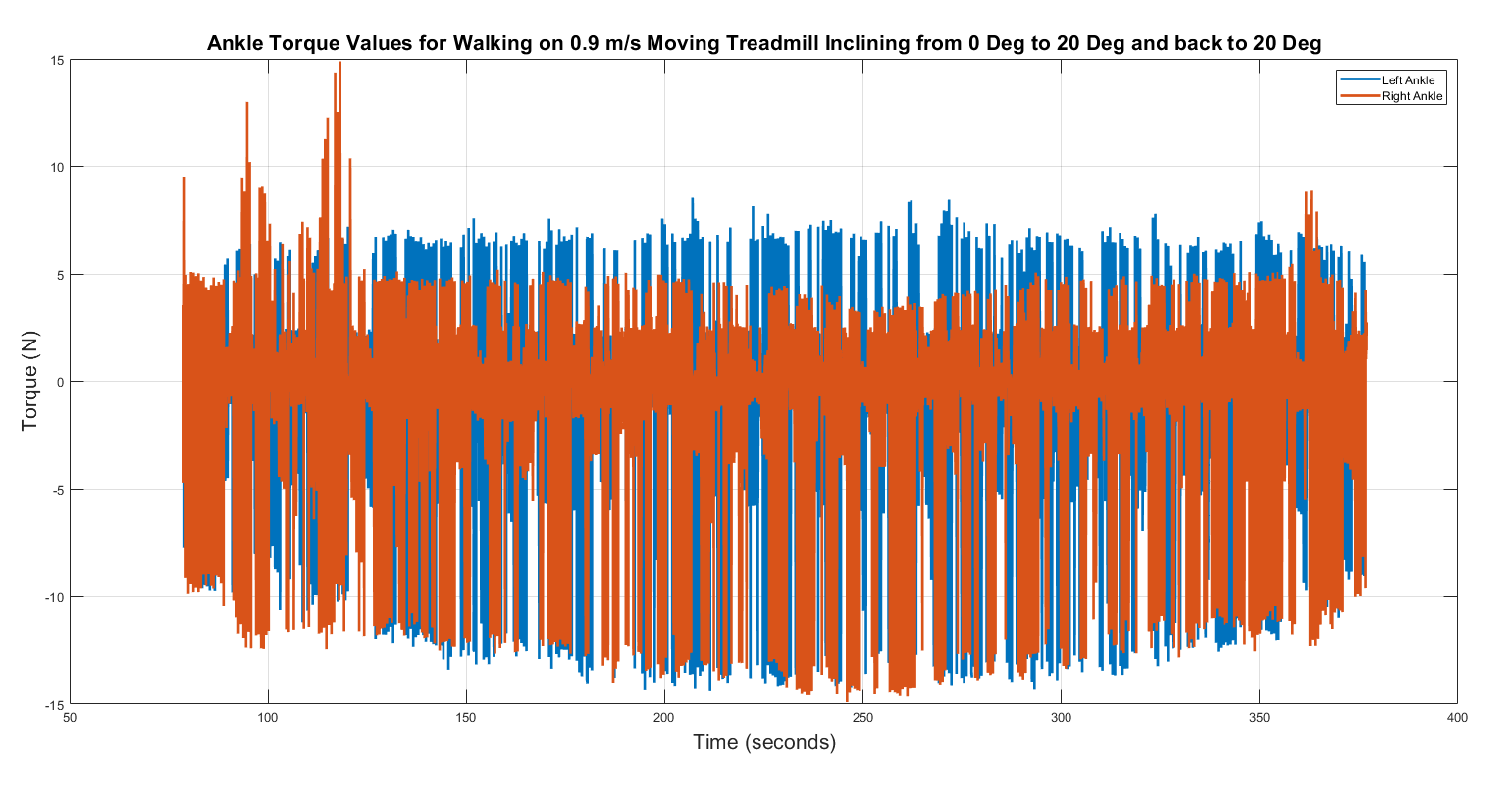}
    \caption{Plot of left and right ankle torque values for hardware experiment on the Cassie bipedal robot walking on an inclined moving treadmill moving at a constant speed of 0.9 m/s. The treadmill is gradually inclined from 0 degrees to its maximum incline of 20 degrees and back to 0 degrees.}
    \label{fig:hardwarePlotFlatTo20ToFlat}
\end{figure}

Figure \ref{fig:hardwarePlotFlatTo20ToFlat} provides insight into the left and right ankle torque values during this experiment. Importantly, the ankle torque values consistently remain well below the maximum and minimum thresholds of $\pm$23 N, as allowed by the Quadratic Program (QP) for the MPC.
\subsection{Transitioning from Stationary Flat Ground to Inclined Moving Treadmill}
\begin{figure}
    \centering
    \includegraphics[width=0.25\textwidth]{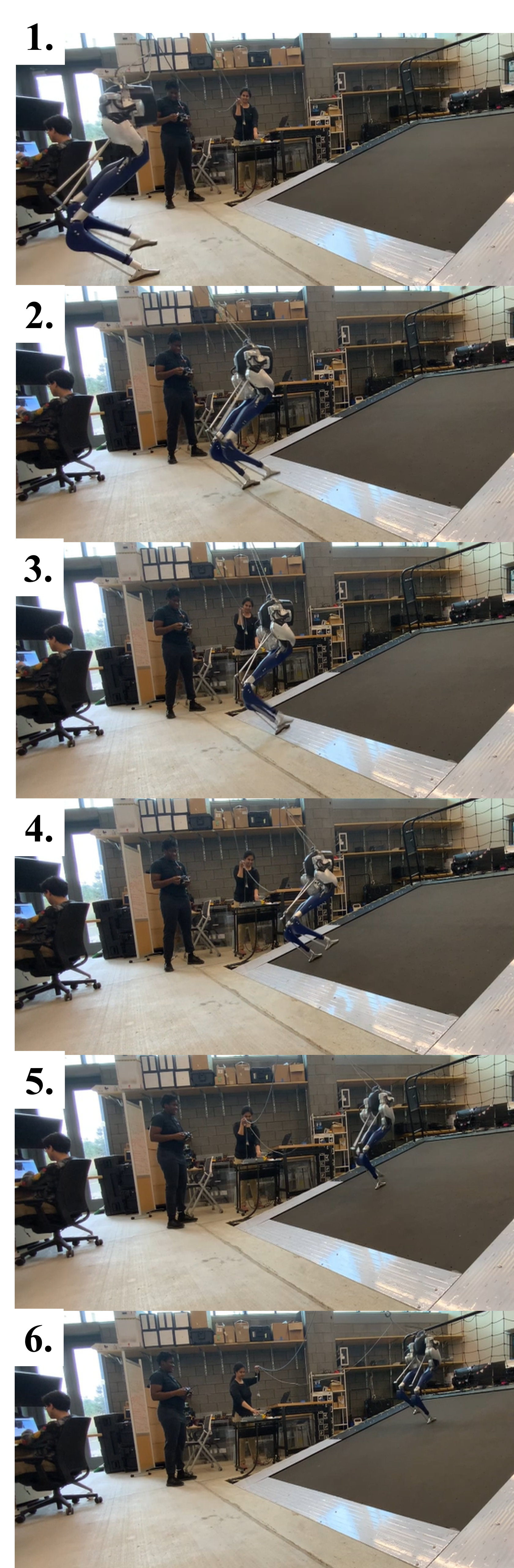}
    \caption{A series of images depicting hardware results of the Cassie bipedal robot walking from flat, stationary ground to a 0.2 m/s moving treadmill inclined at 20 degrees from a back view perspective. Once Cassie steps on the treadmill, the treadmill's speed is gradually increased to 1.5 m/s. The original video can be found in \cite{DynamicLegLocomotion_FlatTo20Deg}.}
    \label{fig:hardwareFlatTo20degFront}
\end{figure}

In our subsequent experiment, we tasked the Cassie bipedal robot with walking from stationary flat ground to an inclined moving treadmill. This challenging scenario was systematically tested for inclined transitions at 4 degrees \cite{DynamicLegLocomotion_FlatTo4Deg}, 8 degrees \cite{DynamicLegLocomotion_FlatTo8Deg}, 15 degrees \cite{DynamicLegLocomotion_FlatTo15Deg}, and 20 degrees \cite{DynamicLegLocomotion_FlatTo20Deg}. Figure \ref{fig:hardwareFlatTo20degFront} shows outtakes for the experiment of the transition from stationary flat ground to a 20-degree incline. These experiments presented two significant sources of disturbance, both of which our controller adeptly surmounted.

The first challenge involved negotiating the large gaps between the stationary ground and the inclined moving treadmill. This transition demanded a seamless adjustment in gait and stability, which our controller effectively managed. The second challenge arose from the fact that the treadmill was in motion at 0.2 m/s while Cassie initiated the transition from stationary flat ground to the treadmill. Despite this differential in speed, Cassie not only successfully navigated the transition but also exhibited the capability to accommodate increased treadmill speeds.

The controller demonstrated its resilience by overcoming both disturbances encountered during the transition process. Furthermore, we successfully escalated the treadmill speed to the maximum values detailed in Section \ref{sec:hardwareMaxWalkingSpeed} post-transition, affirming the adaptability and robust performance of the developed controller in challenging real-world scenarios.

\begin{figure}
    \centering
    \includegraphics[width=0.45\textwidth]{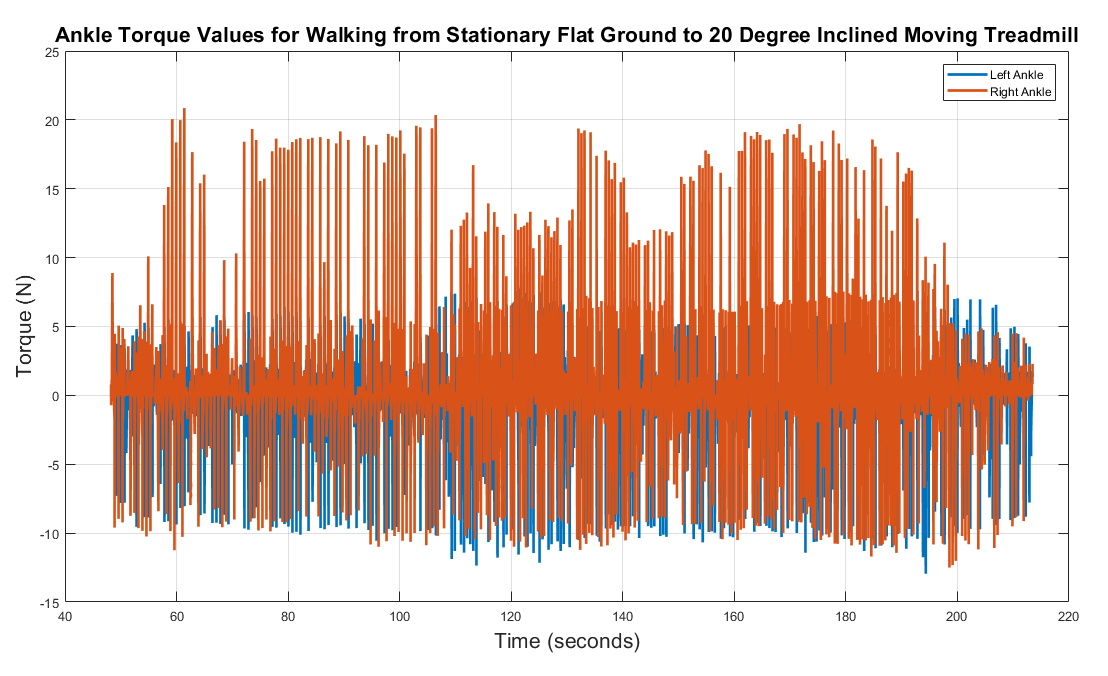}
    \caption{Plot of left and right ankle torque values for hardware experiment on the Cassie bipedal robot walking from stationary flat ground to a 0.2 m/s moving treadmill inclined at 20 degrees. Once Cassie steps on the treadmill, the treadmill's speed is gradually increased to 1.5 m/s.}
    \label{fig:hardwarePlotFlatTo20}
\end{figure}

Figure \ref{fig:hardwarePlotFlatTo20} shows a plot of the values of the left and right ankle torque values during the experiment, from Cassie's transition from flat stationary ground to the 20-degree inclined moving treadmill. While the ankle torque values approach the saturation threshold of $\pm$23 N, as allowed by the QP for the MPC, they remain below this limit, underscoring the controller's effectiveness in managing the complexities of dynamic transitions.

\subsection{Navigating on and off of a Moving Walkway}
In our culminating experiment, we tasked Cassie with traversing on and off a moving walkway—--an essential capability for robots in human-centric environments such as airports and shopping malls. The integration of robots into everyday scenarios necessitates their adept navigation through commonplace features like moving walkways, which typically exhibit average speeds ranging from 0.5 to 0.83 m/s \cite{kusumaningtyas2008accelerating}. Our objective was to validate Cassie's ability to seamlessly adapt to different speeds on the moving walkway. We successfully tested Cassie's performance on moving walkways operating at speeds of 0.5 m/s \cite{DynamicLegLocomotion_05Moving}, 0.8 m/s \cite{DynamicLegLocomotion_08Moving}, and 1.2 m/s \cite{DynamicLegLocomotion_12Moving}.

A notable challenge introduced during this experiment was the presence of a gantry, which did not extend far enough to cover the entire length of the treadmill used to simulate the moving walkway. As a result, Cassie experienced significant tugging forces from the gantry. During the initial setup, when Cassie was walked to the end of the treadmill, it stumbled and nearly fell due to the tugging forces but swiftly recovered \cite{DynamicLegLocomotion_12Moving}. Furthermore, despite operator error leading to Cassie approaching the moving walkway at a diagonal for the 0.8 m/s and 1.2 m/s scenarios, Cassie adeptly navigated onto and off the moving walkway. This demonstration underscores the robustness and adaptability of our controller in addressing real-world challenges during robot locomotion on dynamic surfaces.

\begin{figure}
    \centering
    \includegraphics[width=0.45\textwidth]{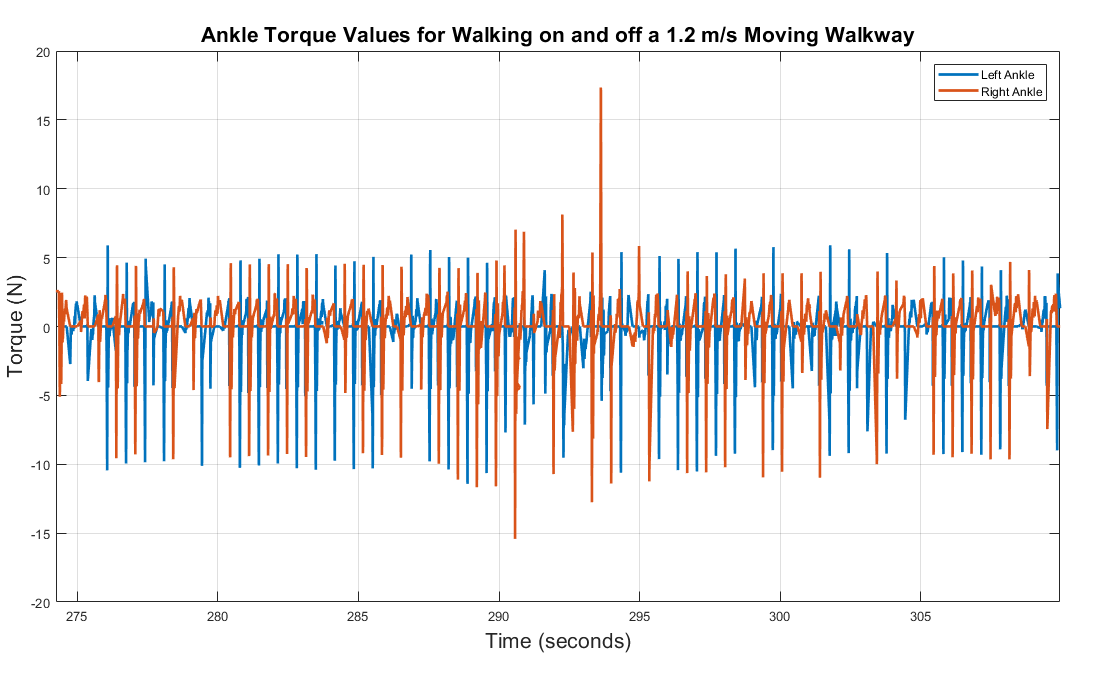}
    \caption{Plot of left and right ankle torque values for hardware experiment on the Cassie bipedal robot walking on to and off of a 1.2 m/s moving walkway (treadmill).}
    \label{fig:hardwarePlot12MovingWalkway}
\end{figure}

Figure \ref{fig:hardwarePlot12MovingWalkway} provides insight into the left and right ankle torque values during Cassie's traversal on and off a 1.2 m/s moving treadmill. Noteworthy spikes are evident between the 290-second and 295-second marks, corresponding to the moment when Cassie stumbled. Importantly, the ankle torque values consistently remain well below the maximum and minimum thresholds of $\pm$23 N, as allowed by the QP for the MPC. This observation attests to the controller's ability to manage unexpected disturbances while maintaining stability, further validating its effectiveness in real-world scenarios.

\section{Conclusions}
This paper details the implementation of controllers on the physical 20 DoF Cassie bipedal robot hardware, moving beyond the commendable SimMechanics simulation outcomes of \cite{dosunmu2023stair}. The transition to physical hardware accentuates the persistent challenge of the sim-to-real gap, encompassing issues like motor friction, sensor inaccuracies, and unforeseen variables absent in simulation environments.

The stringent maximum controller execution time of 2 kHz on Cassie hardware posed a critical concern. While the passivity-based controller (PBC) successfully adheres to this constraint, the model predictive controller (MPC) demands a longer execution time. Our strategic approach involves offloading MPC computations to a secondary computer and further optimizing execution time with CasADi. Successful simulation experiments affirm the viability of this offloading strategy.

Evaluating Cassie's performance with the developed controller structure reveals enhanced walking capabilities. Cassie achieves speeds of up to 2.2 m/s on a flat treadmill, with maintained speeds of 2 m/s on an 8-degree inclined treadmill and 1.5 m/s on a 20-degree incline. This demonstrates the controller's adaptability and improved performance across varying incline levels.

A continuous walking experiment on changing incline further validates the robustness of the controller. Nominal trajectories, not perfectly optimized for all situations, showcase the controller's stability during continuous walking on inclines, emphasizing its adaptability.

Transitions from stationary flat ground to an inclined moving treadmill present unique challenges, including negotiating gaps and adapting to treadmill motion. The controller adeptly manages these disturbances, enabling successful transitions, even at increased treadmill speeds. This resilience underscores the adaptability and robust performance of the developed controller.

Cassie's navigation on and off a moving walkway, despite challenges introduced by a tugging gantry and operator error, highlights the controller's robustness and adaptability in real-world scenarios. This successful demonstration contributes valuable insights for integrating bipedal robots into dynamic human-centric environments.

In conclusion, this paper demonstrates successful implementation and real-world applicability of our novel controller on Cassie hardware. By addressing challenges in execution time, hardware constraints, and adapting to dynamic surfaces, our controller contributes to the advancement of bipedal robots, fostering capabilities for enhanced mobility in diverse environments.

\section*{Acknowledgment}
\small{
 Funding for this work was in part provided by NSF Award No.~2118818. The work of O. Dosunmu-Ogunbi was partially supported by a Rackham Merit Fellowship.}

\nocite{*}
\balance
\bibliographystyle{IEEEtran}
\bibliography{references.bib}

\end{document}